\pdfoutput=1

\documentclass[11pt]{article}

\usepackage{EMNLP2022}
\usepackage{graphicx}
\usepackage{times}
\usepackage{latexsym}
\usepackage{amssymb}
\usepackage[T1]{fontenc}

\usepackage[utf8]{inputenc}

\usepackage{microtype}

\usepackage{inconsolata}
\usepackage{colortbl}
\usepackage{array}
\usepackage{multirow}
\usepackage{ragged2e}
\title{Improving Precancerous Case Characterization via Transformer-based Ensemble Learning}

\author{Yizhen Zhong, Jiajie Xiao, Thomas Vetterli, Mahan Matin,   \\ {\bf Ellen Loo, Jimmy Lin, Richard Bourgon, Ofer Shapira} \\
  Freenome, South San Francisco, CA \\
  \texttt{\{yizhen.zhong, jiajie.xiao, thomas.vetterli, mahan.matin\}@freenome.com}\\
  \texttt{\{ellen.loo, jimmy.lin, richard.bourgon, ofer.shapira\}@freenome.com}}

\begin{document}
\maketitle
\begin{abstract}
The application of natural language processing (NLP) to cancer pathology reports has been focused on detecting cancer cases, largely ignoring precancerous cases. Improving the characterization of precancerous adenomas assists in developing diagnostic tests for early cancer detection and prevention, especially for colorectal cancer (CRC). Here we developed transformer-based deep neural network NLP models to perform the CRC phenotyping, with the goal of extracting precancerous lesion attributes and distinguishing cancer and precancerous cases. We achieved 0.914 macro-F1 scores for classifying patients into negative,  non-advanced adenoma, advanced adenoma and CRC. We further improved the performance to 0.923 using an ensemble of classifiers for cancer status classification and lesion size named entity recognition (NER). Our results demonstrated the potential of using NLP to leverage real-world health record data to facilitate the development of diagnostic tests for early cancer prevention.  

\end{abstract}

\section{Introduction}

Cancer has been the second leading cause of death with more than 1,900k new cases and 600k cancer deaths in the United States in 2022 \citep{siegel2019cancer}. Among those, colorectal cancer (CRC) is the third most common cancer and the third leading cause of cancer death \citep{siegel2019cancer}. Detecting CRC at its early stage can dramatically improve clinical outcomes. The 5-year survival rate is 90\% when colorectal cancer is identified at the localized stage compared to 73\% and 17\% survival rates at the regional or distant stage, respectively\footnote{\url{https://www.cancer.org/cancer/colon-rectal-cancer/detection-diagnosis-staging/survival-rates.html}}. 

CRC progresses from asymptomatic non-advanced adenoma (NAA) to advanced adenoma (AA) and then to invasive carcinoma \citep{junca2020detection}. AAs are adenomas characterized by villous or tubulovillous histology, adenomas or sessile serrated lesions $\geqslant 10$mm, or high-grade dysplasia \cite{junca2020detection, shaukat2021acg}. AA indicates an intermediate or high risk for CRC \cite{lieberman2012guidelines} and requires CRC screening every three years \cite{lieberman2012guidelines}. Recent economic studies suggest a test with increasing adenoma sensitivity in a blood-based CRC screening test can reduce CRC incidence and reduce mortality \cite{putcha2022interception}. 

There is great interest in developing noninvasive diagnostic tests with high sensitivity and specificity for advanced adenoma and CRC screening \cite{imperiale2014multitarget, putcha2022interception}. This development process requires biomarker discovery and clinical validation based on samples collected from large numbers of individuals whose colorectal cancer statuses are confirmed by colonoscopy \cite{putcha2022prevention}. Correctly classifying the colorectal cancer statuses, namely negative (NEG), NAA, AA and CRC, requires expertise in distilling and interpreting tumor stage and histology information and size of precancerous adenoma from colonoscopy and pathology reports. Such nuanced annotations are typically not documented and collected in structured sections of electronic health records or standardized via International Classification of Diseases (ICD) codes \cite{raju2015natural, raju2013adenoma}. Therefore, extracting this information from colonoscopy and pathology reports and generating reliable CRC status classification has heavily relied on manual review by trained gastrointestinal pathologists. Such review is time-consuming, costly and difficult to scale. 

\begin{figure*}[h]
\includegraphics[width=\textwidth]{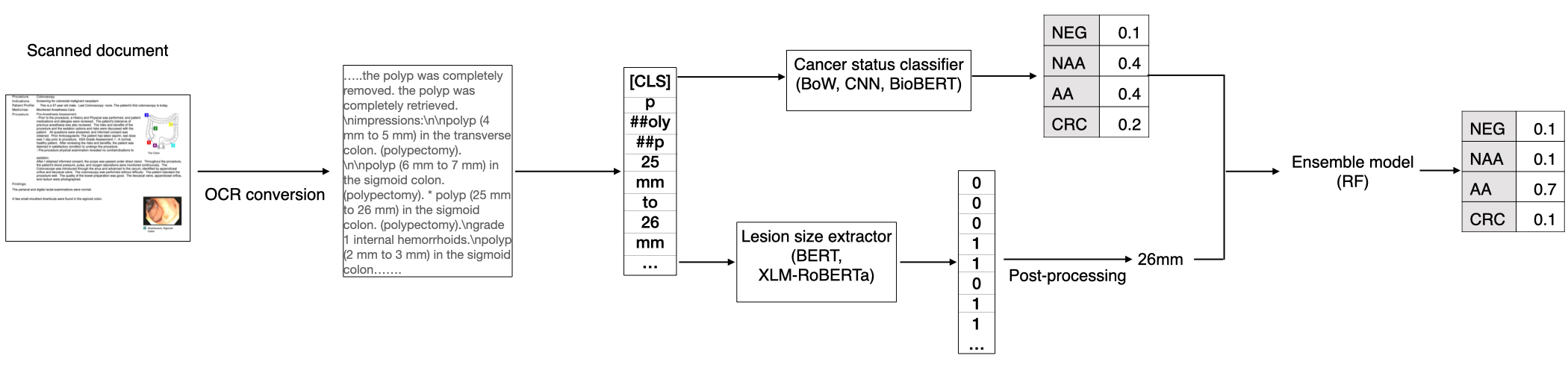}
\caption{Overall framework}\label{figure1}
\end{figure*}

To reduce the burden of manually annotating thousands to hundreds of thousands of pathology reports, and to facilitate the development of noninvasive diagnostic tools for colorectal cancer prevention, we investigated classical and advanced natural language processing (NLP) methods to automatically extract precancerous lesion information and determine CRC status (Figure \ref{figure1}). We developed transformer models to extract both categorical and numerical attributes from colonoscopy and pathology reports. Compared to Bag-of-Word (BoW) and convolutional neural network (CNN) models (see Data and methods in section \ref{data and methods}), we achieved the best performance by fine-tuning the BioBERT model. Since lesion size is an important factor to distinguish between the AA and NAA classes (Appendix \ref{crc_classification}, \citet{winawer2002advanced}), we developed an entity recognition model for lesion size extraction and improved its performance through transfer learning from a non-biomedical domain. We further improved the cancer status classification model performance by explicitly adding extracted lesion size through an ensemble model. Our study also addressed two challenges for NLP practice that are specific to the biomedical industry setting: annotation at the sentence level for numerical variable extraction is limited; and most clinical trial studies that enroll patients from various sites still receive health records in the scanned PDF format \cite{raju2015natural}, creating challenges for precisely locating the diagnosis section in health records. Our research demonstrated that,  along with domain knowledge-informed feature learning, fine-tuned advanced deep learning methods are able to achieve high accuracy in  highly complex and nuanced disease phenotyping tasks, even with only several thousands of documents for model training. 

\section{Related Work}
NLP methods have been applied to pathology reports to extract categorical attributes associated with cancer diagnosis. \citet{yala2017using} used machine learning methods with Bag-of-Words features to classify patients into breast cancer carcinoma and atypia categories. Adding clinical concepts from the Unified Medical Language System (UMLS) was shown to improve classification performance \cite{li2010information, martinez2011information}.  \citet{preston2022towards} used embedding vectors that are pre-trained in BERT-based \cite{devlin2018bert} models for tumor site, histology and TNM staging (T: tumor size/location; N: lymph node status; M: metastasis) classification from longitudinal reports and developed classifiers to detect cancer cases. \citet{park2021improving} extracted cancer histology, site and surgical procedure from colon, lung and kidney cancer data. They demonstrated good performance by leveraging transfer learning across cancer types and few-shot learning by accounting for semantic similarity. Other deep learning approaches such as hierarchical attention neural networks \cite{gao2018hierarchical, gao2019classifying}, multitask learning \cite{alawad2020automatic}, and graph convolutional networks \cite{wu2020structured} have been employed to extract cancer characteristics such as primary site and histological grade. However, these studies primarily focused on extracting categorical cancer characteristics that are routinely collected in cancer registries \cite{klein2011standards} and largely ignored numerical and precancerous attributes, which are critical for developing early cancer detection technology.

The extraction of numerical cancer attributes is challenging because the semantic context for numerical variables is mostly at the  sentence level instead of document/patient level \cite{li2010information, odisho2020natural}. This creates a discrepancy between training objectives (sentence level) and output evaluation (patient level). To overcome this limitation, \citet{li2010information} first identified the sentences that contain the numerical values and extracted them through regular expression matching. \citet{aalabdulsalam2018automated} treated TNM staging extraction as a sequence labeling task with pattern matching and conditional random field techniques. \citet{odisho2020natural} encoded tokens and their context words as bag-of-\emph{n}-gram features and classified the token sequence for TNM staging and tumor volume extraction. In our work, we treated numerical lesion size extraction as a named entity recognition (NER) task and addressed the challenge of limited annotated sample size by transfer learning from models pre-trained in a non-biomedical domain. 

Previous attempts to employ NLP methods to parse colonoscopy reports and linked pathology reports have aimed to characterize adenomas due to their importance in estimating colorectal cancer risk \cite{lee2019accurate, raju2013adenoma, raju2015natural, imler2013natural}. The limitations of current studies are two fold: first, most studies still rely on rule-based systems such as Linguamatics \cite{lee2019accurate} and cTAKEs \cite{imler2013natural, savova2010mayo} for extracting adenomas through pattern-matching and dictionary look-up. Deriving rules can be time-consuming, require extensive domain knowledge, and likely results in overfitting to the development dataset and limited portability; second, NLP studies for colorectal cancer do not perform end-to-end CRC phenotyping to classify patients into NEG, NAA, AA and CRC, which are of great interest to characterize CRC risk and prioritize patients for cancer screening and early cancer detection. Here we performed end-to-end precancerous and cancer status characterization with deep learning methods which promise to be more generalizable and efficient than rule-based approaches.

\section{Data and methods}\label{data and methods}
\subsection{Dataset}
In this study, we used health records from 3,068 patients collected as part of two studies from 68 collection sites. In some cases, multiple types of health records are associated with one patient, including colonoscopy, pathology, surgical pathology, and radiology reports. In total, there are 5,405 documents for all patients. Appendix Figure \ref{document number distribution} shows the distribution of document numbers for each patient.

We split patients into train and test sets stratified by cancer status. To assess the generalizability of NLP models when applied to pathology reports collected in unseen sites, we used samples from independent collection sites for the train and test sets. There are 2,149 samples in the training set from 54 sites and 919 samples in the test set from 14 independent sites. Appendix Table \ref{sample_counts} shows the sample count for each cancer status in the train and test sets.

\subsection{Document and sentence level annotations}
A certified pathologist reviewed and assigned patient-level labels for colorectal cancer status and lesion size. The detailed annotation criteria are described in Appendix \ref{crc_classification}. For lesion size annotation, the pathologist first identified the index lesion, which is the most clinically significant lesion according to cancer status and lesion type. Then the size of the index lesion was used as the patient-level lesion size annotation. We used zero as the lesion size for healthy samples with no identified lesions. 

We generated sentence-level annotation for lesion size. We treated the lesion size named entity annotation as a binary label with tokens within the lesion size entity as 1 and tokens outside the lesion size entity as 0. We did not distinguish the start or end token of the named entity. We randomly selected 499 documents from 225 patients from the training set and 331 documents from 114 patients from the test set for NER model training and evaluation, respectively.

\subsection{Data preprocessing}
Since the reports are in scanned PDF format, we first digitized the reports with optical character recognition provided by the Google Vision API\footnote{\url{https://cloud.google.com/vision/docs/ocr}}. The OCR algorithm outputs the recognized text and the coordinates of the bounding box for each text block. 

We used fuzzy matching for words: "diagnosis," "finding," "impression," "diagnoses," "findings,", "impressions," and "polyp" with text in each bounding box. This allowed us to identify sections important for diagnosis and ignore irrelevant sections to increase the signal-to-noise ratio. To allow for some error in bounding box identification, we retained texts within 10 bounding boxes and at most 100 words after the first matched bounding box. These keywords for fuzzy matching and window size were determined by an iterative manual inspection of reports in the training set. 

For CNN and BoW, we concatenated documents corresponding to one patient into one text segment and padded the concatenated text segment to the max length. For the BERT models with a sequence length limit of 512 tokens, we split the text into segments with 10 overlapping tokens.  

\subsection{Cancer status classification model}
We built and tested three models, namely Bag-of-Words (BoW) \cite{zhong2018characterizing}, CNN \cite{kim2019convolutional} and BioBERT \cite{lee2020biobert}, for cancer status classification. We split the 2,149 documents reserved for training into training and validation sets in a 9:1 ratio and selected the best model based on the validation macro-F1 score [See \ref{metrics}].
For BoW, \emph{tf-idf} representation \cite{zhong2018characterizing}, a term-frequency based featurization, was derived as input features for SVM models with linear, polynomial or radial basis function (RBF) kernels. We used unigram features and removed terms that appear in less than 10 documents. 

For the CNN model, instead of doing hand-crafted feature engineering, 1D convolution kernels were learned to extract localized text patterns from pathology reports. The convolutional layer is followed by a max-pooling layer and a fully connected layer to classify colorectal cancer status. 

For the BERT model, BioBERT \cite{lee2020biobert}, a pre-trained biomedical language representation, was employed and fine-tuned as follows to encode the pathology reports for cancer status classification. BioBERT was pre-trained based on BERT initiated weights with biomedical domain corpora (PubMed abstracts and PMC full-text articles) and has increased performance in biomedical text mining tasks including NER, relation extraction and question-answering. We added a fully connected layer after the [CLS] embedding vector for multiclass classification \cite{devlin2018bert}. Because one patient can be associated with multiple documents or text segments but the cancer status label is annotated for each patient, this creates a multiple instance learning problem. We treated each patient as a bag and each text segment as an instance within the bag. We used max-pooling to get the largest softmax probability for each class across multiple text segments and renormalize with the softmax function to calculate cross-entropy loss per patient.

All these models were optimized through Bayesian hyperparameter tuning \cite{snoek2012practical, shahriari2015taking} with early stopping from the Hyperband algorithm \cite{li2017hyperband}. More details of the procedures can be found in the Appendix Table \ref{hpo}.

\subsection{Lesion size extraction model}
We treated the lesion size extraction as a NER task and compared two approaches. For direct fine-tuning, we used a pretrained BERT-base-uncased\footnote{\url{https://huggingface.co/bert-base-uncased}} model and classified token embedding into binary labels where the positive label indicates the target named entity. We also fine-tuned a XLM\_RoBERTa\_base\footnote{\url{https://huggingface.co/xlm-roberta-base}} model.

Observing the similarity between lesion size vs. one of the annotated named entities (QUANTITY: Measurements, as of weight or distance) from the OntoNotes5 corpus \cite{weischedel2011ontonotes}, we used an XLM\_RoBERTa\footnote{\url{https://huggingface.co/asahi417/tner-xlm-roberta-base-uncased-ontonotes5}} model that was previously fine-tuned on the OntoNotes5 dataset for NER of QUANTITY. We then continued to fine-tune this model on the cancer pathology dataset for lesion size extraction to explore the benefit of transfer learning.
 Both direct fine-tuning and transfer learning models were trained and validated based on a 7:1 split of the sentence-level annotated documents. We selected the model with the best validation F1 score and evaluated its performance on the holdout test set. The hyperpameters can be found in Appendix Table \ref{hpo}. 
\subsection{Ensemble model}
We built an ensemble model with BioBERT predicted probability for each class and binarized lesion size feature (lesion size $\geqslant10$mm or not). For the model training, we used the binarized ground-truth lesion size. For model inference, we used the binarized NER-extracted lesion size. We trained a random forest model with 10 trees and max\_depth=10 as the ensemble model on the training set and tested its performance on the validation and test sets for the cancer status classification task.
\subsection{Metrics}\label{metrics}
For cancer status classification evaluation, we computed the precision, recall and F1 for each cancer status. We used the macro-F1=$\sum_i^n\frac{F_i}{n}$ for the overall model performance metric for multi-class classification ($n$ classes).  

For NER evaluation, we count consecutive positive labeled tokens as one named entity. We identified named entities derived from ground-truth labels (total ground-truth positive, TGP) and predicted labels (total predicted positive, TPP). We counted an exact match of starting and ending index of the ground-truth and predicted entities as true positive (TP). We calculated the precision as $\frac{TP}{TPP}$, recall as $\frac{TP}{TGP}$ and F1 score for the named entity recognition task.

\begin{table*}[h]
\centering
\small
\resizebox{\textwidth}{!}{
\begin{tabular}{>{\centering\hspace{0pt}}m{0.317\linewidth}>{\centering\hspace{0pt}}m{0.119\linewidth}>{\centering\hspace{0pt}}m{0.113\linewidth}>{\centering\hspace{0pt}}m{0.113\linewidth}>{\centering\hspace{0pt}}m{0.119\linewidth}>{\centering\arraybackslash\hspace{0pt}}m{0.146\linewidth}} 
\hline
\textbf{Model} & \textcolor[rgb]{0.114,0.11,0.114}{\textbf{NEG F1} }\par{}\textcolor[rgb]{0.114,0.11,0.114}{(n=515)} & \textcolor[rgb]{0.114,0.11,0.114}{\textbf{NAA F1} }\par{}\textcolor[rgb]{0.114,0.11,0.114}{(n=183)} & \textcolor[rgb]{0.114,0.11,0.114}{\textbf{AA F1} }\par{}\textcolor[rgb]{0.114,0.11,0.114}{(n=146)} & \textcolor[rgb]{0.114,0.11,0.114}{\textbf{CRC F1} }\par{}\textcolor[rgb]{0.114,0.11,0.114}{(n=75)} & \textcolor[rgb]{0.114,0.11,0.114}{\textbf{Macro-F1}} \\ 
\hline
\textcolor[rgb]{0.114,0.11,0.114}{\textbf{BoW}} & \textcolor[rgb]{0.114,0.11,0.114}{0.941} & \textcolor[rgb]{0.114,0.11,0.114}{0.706} & \textcolor[rgb]{0.114,0.11,0.114}{0.758} & \textcolor[rgb]{0.114,0.11,0.114}{0.952} & \textcolor[rgb]{0.114,0.11,0.114}{0.839} \\
\textcolor[rgb]{0.114,0.11,0.114}{\textbf{CNN}} & \textcolor[rgb]{0.114,0.11,0.114}{0.971} & \textcolor[rgb]{0.114,0.11,0.114}{0.842} & \textcolor[rgb]{0.114,0.11,0.114}{0.773} & \textcolor[rgb]{0.114,0.11,0.114}{0.947} & \textcolor[rgb]{0.114,0.11,0.114}{0.883} \\
\textcolor[rgb]{0.114,0.11,0.114}{\textbf{BioBERT}} & \textcolor[rgb]{0.114,0.11,0.114}{\textbf{0.972}} & \textcolor[rgb]{0.114,0.11,0.114}{0.888} & \textcolor[rgb]{0.114,0.11,0.114}{0.833} & \textcolor[rgb]{0.114,0.11,0.114}{0.962} & \textcolor[rgb]{0.114,0.11,0.114}{0.914} \\
\textcolor[rgb]{0.114,0.11,0.114}{\textbf{BioBERT+Lesion Size}} & \textcolor[rgb]{0.114,0.11,0.114}{0.965} & \textcolor[rgb]{0.114,0.11,0.114}{\textbf{0.894}} & \textcolor[rgb]{0.114,0.11,0.114}{\textbf{0.854}} & \textcolor[rgb]{0.114,0.11,0.114}{\textbf{0.980}} & \textcolor[rgb]{0.114,0.11,0.114}{\textbf{0.923}} \\
\hline
\end{tabular}}
\caption{\label{test_classification_by_cancer_type}Performance of cancer status classifiers as measured by the test F1 scores for each class and their macro-averages.}
\end{table*}

\section{Results}
\subsection{Cancer status classification model performance}
We treated the cancer status (CRC, AA, NAA and NEG) extraction as a document classification problem and trained BoW, CNN and BioBERT models. All models achieved over 0.8 macro-F1 scores, with the BioBERT model outperforming BoW and CNN (Table \ref{test_classification_by_cancer_type}). In particular, all models including BoW classified CRC and NEG with high accuracy (> 0.9 F1 score) and AA and NAA with lower accuracy (< 0.8 F1 score). This suggests that unigram features are sensitive for classifying cancer and healthy patients from pathology reports but less sensitive for differentiating precancerous patients. 

The CNN model (NAA F1=0.842, AA F1=0.773) improved NAA and AA performance compared to BoW (NAA F1=0.706, AA F1=0.758).  This suggests that the larger kernels used in CNN improve the capture of semantics for precancerous classes compared with unigram features. The BioBERT model further improved AA and NAA performance (NAA F1=0.888, AA F1=0.833). As the model complexity and its ability to capture long-range interaction increases, the model performed better. Although the number of training samples was limited, the more complex models appear to be more generalizable.

\begin{table}[h]
\centering
\small
\resizebox{\linewidth}{!}{%
\begin{tabular}{>{\centering\hspace{0pt}}m{0.233\linewidth}>{\centering\hspace{0pt}}m{0.113\linewidth}|>{\centering\hspace{0pt}}m{0.113\linewidth}>{\centering\hspace{0pt}}m{0.108\linewidth}>{\centering\hspace{0pt}}m{0.092\linewidth}>{\centering\hspace{0pt}}m{0.113\linewidth}>{\centering\arraybackslash\hspace{0pt}}m{0.115\linewidth}} 
\cline{3-7}
\multicolumn{2}{>{\Centering\hspace{0pt}}m{0.346\linewidth}|}{\multirow{2}{*}{\Centering{}}} & \multicolumn{4}{>{\Centering\hspace{0pt}}m{0.426\linewidth}}{\textcolor[rgb]{0.114,0.11,0.114}{\textbf{Predicted Label}}} &  \\
\multicolumn{2}{>{\Centering\hspace{0pt}}m{0.346\linewidth}|}{} & \textcolor[rgb]{0.114,0.11,0.114}{NEG} & \textcolor[rgb]{0.114,0.11,0.114}{NAA} & \textcolor[rgb]{0.114,0.11,0.114}{AA} & \textcolor[rgb]{0.114,0.11,0.114}{CRC} & \textcolor[rgb]{0.114,0.11,0.114}{\textbf{}{Total}} \\ 
\hline
\multirow{4}{*}{\Centering{}\textcolor[rgb]{0.114,0.11,0.114}{\textbf{True Label}}} & \textcolor[rgb]{0.114,0.11,0.114}{NEG} & \textcolor[rgb]{0.114,0.11,0.114}{1195} & \textcolor[rgb]{0.114,0.11,0.114}{8} & \textcolor[rgb]{0.114,0.11,0.114}{15} & \textcolor[rgb]{0.114,0.11,0.114}{0} & \textcolor[rgb]{0.114,0.11,0.114}{1218} \\
 & \textcolor[rgb]{0.114,0.11,0.114}{NAA} & \textcolor[rgb]{0.114,0.11,0.114}{7} & \textcolor[rgb]{0.114,0.11,0.114}{456} & \textcolor[rgb]{0.114,0.11,0.114}{17} & \textcolor[rgb]{0.114,0.11,0.114}{1} & \textcolor[rgb]{0.114,0.11,0.114}{481} \\
 & \textcolor[rgb]{0.114,0.11,0.114}{AA} & \textcolor[rgb]{0.114,0.11,0.114}{5} & \textcolor[rgb]{0.114,0.11,0.114}{13} & \textcolor[rgb]{0.114,0.11,0.114}{257} & \textcolor[rgb]{0.114,0.11,0.114}{2} & \textcolor[rgb]{0.114,0.11,0.114}{277} \\
 & \textcolor[rgb]{0.114,0.11,0.114}{CRC} & \textcolor[rgb]{0.114,0.11,0.114}{1} & \textcolor[rgb]{0.114,0.11,0.114}{0} & \textcolor[rgb]{0.114,0.11,0.114}{0} & \textcolor[rgb]{0.114,0.11,0.114}{172} & \textcolor[rgb]{0.114,0.11,0.114}{173} \\
\hline
\end{tabular}
}
\caption{\label{error_analysis}
Error analysis for the BioBERT model in the training and validation data.}
\end{table}

\subsection{Lesion size extraction model performance}
We performed an error analysis on the training and validation data to identify the source of incorrect predictions (Table \ref{error_analysis}, Appendix Table \ref{train_val_classification_performance}) for the BioBERT model. We found misclassifications were usually confusion between AA and NAA. 68.0\% (17/25) of incorrect predictions for NAAs were classified as AA, and 65.0\% (13/20) of incorrect predictions for AAs were classified as NAAs. Since lesion size is an important factor to distinguish between the AA and NAA classes (Appendix \ref{crc_classification}), we proposed to explicitly add lesion size as an additional feature to improve the BERT-based cancer status classification model.

The direct fine-tuning of the BERT model for lesion size NER had low performance, potentially due to the small sample size for sentence-level annotation (test F1 score=0.202, precision=0.159 and recall=0.273, Table \ref{NER performance}). We then evaluated the transfer learning approach, using an XLM\_RoBERTa model that had been fine-tuned on the OntoNotes dataset for QUANTITY extraction. Directly applying this model to the cancer pathology dataset to extract lesion size led to an increased F1 (0.508) score, with high recall (0.703) and low precision (0.398). We next continued to train this model on the cancer pathology dataset to perform lesion size extraction. Interestingly, additional fine-tuning substantially improved the performance (F1=0.757, precision=0.761 and recall=0.753), especially the precision. This suggests that transfer learning using models fine-tuned on tasks outside the biomedical domain can substantially improve domain-specific NLP performance, even with a relatively small sample size.

\begin{table*}
\centering

\small
\resizebox{\linewidth}{!}{%
\begin{tabular}{cccccccccc} 
\hline
\multirow{2}{*}{\textbf{Model}} & \multicolumn{3}{c}{\textcolor[rgb]{0.114,0.11,0.114}{\textbf{Train}}} & \multicolumn{3}{c}{\textcolor[rgb]{0.114,0.11,0.114}{\textbf{Val}}} & \multicolumn{3}{c}{\textcolor[rgb]{0.114,0.11,0.114}{\textbf{Test}}} \\
 & \textcolor[rgb]{0.114,0.11,0.114}{\textbf{F1}} & \textcolor[rgb]{0.114,0.11,0.114}{\textbf{precision}} & \textcolor[rgb]{0.114,0.11,0.114}{\textbf{recall}} & \textcolor[rgb]{0.114,0.11,0.114}{\textbf{F1}} & \textcolor[rgb]{0.114,0.11,0.114}{\textbf{precision}} & \textcolor[rgb]{0.114,0.11,0.114}{\textbf{recall}} & \textcolor[rgb]{0.114,0.11,0.114}{\textbf{F1}} & \textcolor[rgb]{0.114,0.11,0.114}{\textbf{precision}} & \textcolor[rgb]{0.114,0.11,0.114}{\textbf{recall}} \\ 
\hline
\textcolor[rgb]{0.114,0.11,0.114}{\textbf{FT\_BERT}} & \textcolor[rgb]{0.114,0.11,0.114}{0.316} & \textcolor[rgb]{0.114,0.11,0.114}{0.240} & \textcolor[rgb]{0.114,0.11,0.114}{0.450} & \textcolor[rgb]{0.114,0.11,0.114}{0.174} & \textcolor[rgb]{0.114,0.11,0.114}{0.142} & \textcolor[rgb]{0.114,0.11,0.114}{0.225} & \textcolor[rgb]{0.114,0.11,0.114}{0.202} & \textcolor[rgb]{0.114,0.11,0.114}{0.159} & \textcolor[rgb]{0.114,0.11,0.114}{0.273} \\
\textcolor[rgb]{0.114,0.11,0.114}{\textbf{FT\_XLM\_RoBERTa}} & \textcolor[rgb]{0.114,0.11,0.114}{0.471} & \textcolor[rgb]{0.114,0.11,0.114}{0.372} & \textcolor[rgb]{0.114,0.11,0.114}{0.644} & \textcolor[rgb]{0.114,0.11,0.114}{0.259} & \textcolor[rgb]{0.114,0.11,0.114}{0.197} & \textcolor[rgb]{0.114,0.11,0.114}{0278} & \textcolor[rgb]{0.114,0.11,0.114}{0.243} & \textcolor[rgb]{0.114,0.11,0.114}{0.186} & \textcolor[rgb]{0.114,0.11,0.114}{0.351} \\
\textcolor[rgb]{0.114,0.11,0.114}{\textbf{OntoNotes\_XLM\_RoBERTa}} & \textcolor[rgb]{0.114,0.11,0.114}{0.395} & \textcolor[rgb]{0.114,0.11,0.114}{0.275} & \textcolor[rgb]{0.114,0.11,0.114}{0.702} & \textcolor[rgb]{0.114,0.11,0.114}{0.360} & \textcolor[rgb]{0.114,0.11,0.114}{0.243} & \textcolor[rgb]{0.114,0.11,0.114}{0.695} & \textcolor[rgb]{0.114,0.11,0.114}{0.508} & \textcolor[rgb]{0.114,0.11,0.114}{0.398} & \textcolor[rgb]{0.114,0.11,0.114}{0.703} \\
\textcolor[rgb]{0.114,0.11,0.114}{\textbf{TL\_OntoNotes\_XLM\_RoBERTa}} & \textcolor[rgb]{0.114,0.11,0.114}{\textbf{0.933}} & \textcolor[rgb]{0.114,0.11,0.114}{\textbf{0.911}} & \textcolor[rgb]{0.114,0.11,0.114}{\textbf{0.956}} & \textcolor[rgb]{0.114,0.11,0.114}{\textbf{0.874}} & \textcolor[rgb]{0.114,0.11,0.114}{\textbf{0.856}} & \textcolor[rgb]{0.114,0.11,0.114}{\textbf{0.893}} & \textcolor[rgb]{0.114,0.11,0.114}{\textbf{0.757}} & \textcolor[rgb]{0.114,0.11,0.114}{\textbf{0.761}} & \textcolor[rgb]{0.114,0.11,0.114}{\textbf{0.753}} \\
\hline
\end{tabular}
}
\caption{\label{NER performance}NER model performance. FT\_BERT: direct fine-tuned BERT model. FT\_XLM\_RoBERTa: direct fine-tuned XLM\_RoBERTa model. OntoNotes\_XLM\_RoBERTa: XLM\_RoBERTa model that has been fine-tuned on OntoNotes dataset. TL\_OntoNotes\_XLM\_RoBERTa: XLM\_RoBERTa model that has been fine-tuned on OntoNotes dataset and pathology lesion size dataset.}

\end{table*}

\begin{figure*}[h]
\raggedleft
\includegraphics[width=\textwidth]{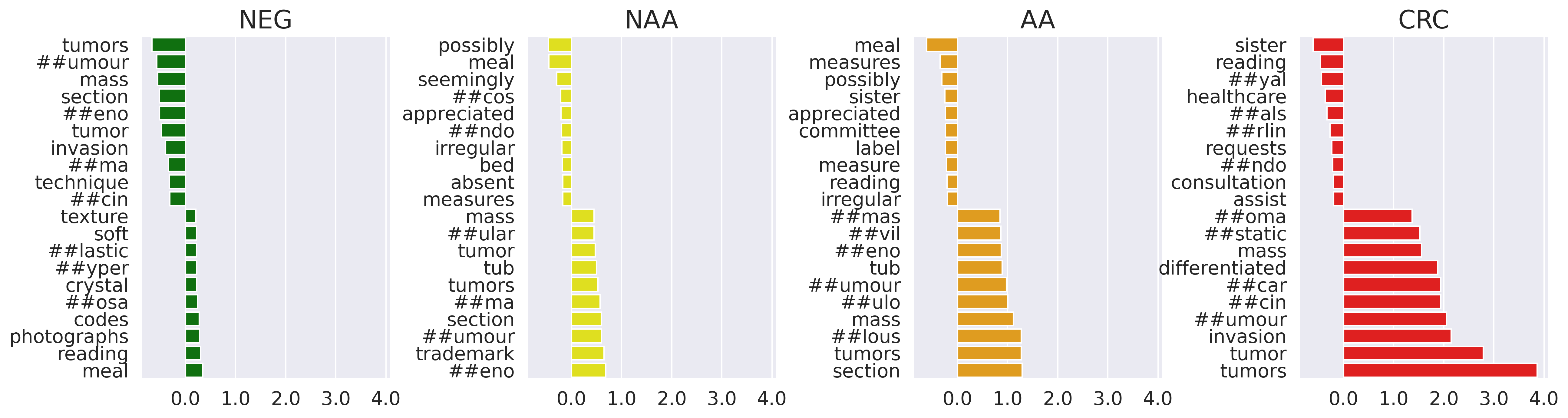}
\caption{Attribution score. Tokens with the largest integrated gradient analysis attribution score for each cancer status. (Top 10 positive and negative attribution scores.)}
\label{ig}
\end{figure*}

\subsection{Ensemble model improves cancer status classification}
We then assessed the effect of explicitly adding lesion size as an additional feature to classify cancer status. The ensemble model which combined BioBERT predicted probabilities and binarized lesion size ($\geqslant 10$mm or not) improved NAA performance from 0.888 to 0.894 and AA performance from 0.833 to 0.854 while maintaining the already high performance for the NEG and CRC classes (Table \ref{test_classification_by_cancer_type}). The macro-F1 was 0.923 for the ensemble model compared to 0.914 for the BioBERT model alone. This suggests that explicitly adding features that are informed by domain knowledge can improve classification performance compared to fine-tuned transformer models alone. This approach may be particularly beneficial for applications in which training data are limited.

\subsection{Integrated gradient analysis}
To investigate which features are most important for BioBERT model performance, we performed integrated gradient analysis, which computes attribution scores that measure feature importance with respect to the classification prediction \cite{sundararajan2017axiomatic}. We calculated attribution with respect to the input embedding vector. We performed integrated gradient analysis for a random subset of 534 NEGs, 197 NAAs, 168 AAs and 80 CRCs and calculated the averaged feature attributions across documents for each class (Figure \ref{ig}). High-scoring tokens related to CRC classification included “tumor,” “invasion,” and “carcinoma.” High-scoring tokens related to AA and NAA classification included  “tub”, “\#\#umour”, and “\#\#eno.” This model interpretability analysis helped to confirm that our NLP model is able to leverage key terms that match domain knowledge.

\section{Conclusion}
Determining cancer status and characterizing precancerous lesions are critical and time-consuming steps for the development and evaluation of diagnostic tests for colorectal cancer screening. Here we achieved a 0.914 macro-F1 score for cancer status classification with transformer models fine-tuned using BioBERT. Informed by the domain knowledge and error analyses, we identified lesion size as a critical factor for differentiating between AAs and NAAs, but one that was not efficiently captured in BioBERT context-dependent embeddings. Using an ensemble model combining a fine-tuned BioBERT model and a lesion size named entity recognition model, we further improved classification performance to a macro-F1 score of 0.923. The lesion size extraction model was developed through transfer learning, using a transformer model trained in a non-biomedical domain. We showed that directly fine-tuning of transformer models was inadequate for domain-specific NLP tasks, and that precise feature engineering and use of ensemble models was needed to improve classification performance. Overall, we provided an accurate algorithm for characterizing precancerous cases that may help to improve early colorectal cancer detection and prevention, and a model training framework that leverages advanced NLP techniques to address complex disease phenotyping tasks in biomedical domain.

\section{Limitations}
One limitation of this work is that we could not fully evaluate how use of scanned reports and OCR affects performance as compared to use of electronic reports, due to a lack of dataset with paired scanned and electronic formats. The scanned format makes the selection of relevant sections from the colonoscopy and pathology reports challenging. We used fuzzy matching of selected keywords to identify sections that are likely important for diagnosis, but this process might introduce bias. Additionally, the digitization process by OCR results in errors in keywords and numerical values. For example, we observed “tubulovillous” was misrecognized as “tubulovillaus” and "0.2cm" is misrecognized as "0:2cm". This could affect the performance in NER and final cancer status classification. Future work could include evaluating other OCR tools besides Google Vision. 

Another limitation is the lack of validation studies using an external dataset. It is known that health records vary substantially in both formats and content. Studies have been done to transform pathology reports to use standardized terminologies and diagnoses \cite{kim2020standardized, ryu2020transformation}. Even though the dataset used in this study is collected from 68 collection sites across the US, the sample size is still relatively small and may not fully capture the variabilities of real-world data.

\section{Acknowledgments} 

This work was supported by Freenome. We thank Paul Tittel and Michael Widrich for helpful discussion; Chuanbo Xu for assistance in acquiring the clinical documents; Amit Pasupathy for helping to collect the reports; and Anooj Patel and David Liu for support with the computational and machine learning infrastructure.

\bibliography{anthology,custom}
\bibliographystyle{acl_natbib}

\appendix

\section{Appendix}
\label{sec:appendix}
\setcounter{table}{0}
\renewcommand{\thetable}{A\arabic{table}}
\setcounter{figure}{0}
\renewcommand{\thefigure}{A\arabic{figure}}
\subsection{CRC status annotation criteria}
\label{crc_classification}
\begin{itemize}
    \item CRC
    \begin{itemize}
        \item All stages (I-IV)
    \end{itemize}
    \item Advanced adenoma (AA)
    \begin{itemize}
        \item Adenoma with carcinoma \emph{in situ} or high-grade dysplasia, any size
        \item Adenoma, any villous features, any size 
        \item Adenoma $\geqslant 1.0$ cm in size
        \item Serrated lesion, $\geqslant 1.0$ cm in size, including sessile serrated adenoma/polyp (SSA/P) with or without cytological dysplasia and hyperplastic polyps (HP) $\geqslant 1.0$ cm
        \item Traditional serrated adenoma (TSA), any size
    \end{itemize}
    \item Non-advanced adenoma (NAA)
    \begin{itemize}
        \item Any number of adenomas, all < 1.0 cm in size, non-advanced
    \end{itemize}
    \item Negative (NEG)
    \begin{itemize}
        \item All SSA/P < 1.0 cm and HP < 1.0 cm NOT in sigmoid or rectum
        \item HP < 1.0 cm in the sigmoid or rectum
        \item Negative upon histopathological review
        \item No findings on colonoscopy, no histopathological review
    \end{itemize}
    
\end{itemize}

\begin{figure*}[ht!]
\includegraphics[width=0.7\linewidth]{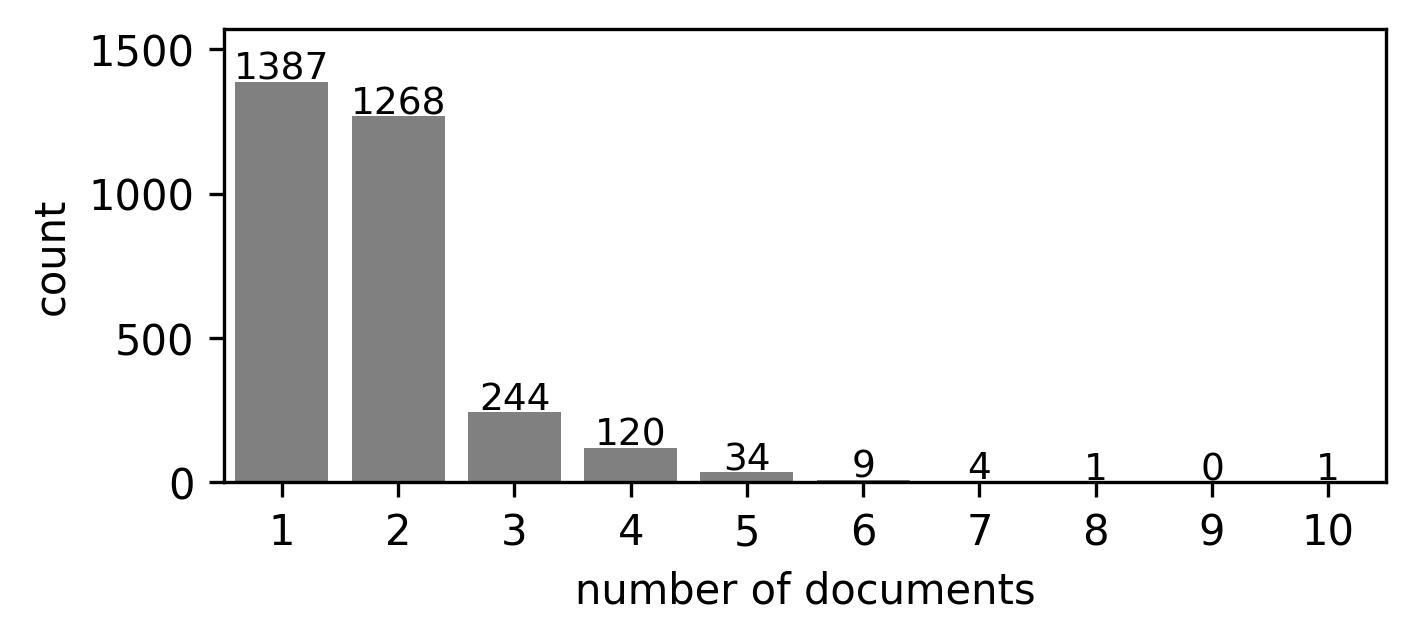}
\centering
\caption{\label{document number distribution}Distribution of document number per patient.}
\label{doc_hist}
\end{figure*}

\begin{table*}[h]
\centering
\setlength{\tabcolsep}{1\tabcolsep}
\small
\begin{tabular}{lccccc}
\hline
\textbf{Sample Count} & \textbf{NEG} & \textbf{NAA} & \textbf{AA} & \textbf{CRC} & \textbf{Total}\\
\hline
\textbf{Train+Val set} &
\small\verb|1,221| & \small\verb|482|& \small\verb|273| & \small\verb|173|& \small\verb|2,149| \\

\textbf{Test set} & \small\verb|515| & \small\verb|183|& \small\verb|146| & \small\verb|75|& \small\verb|919| \\
\hline
\end{tabular}
\caption{\label{sample_counts}Sample count for combined training and validation set and for test set for cancer status classification model.}

\end{table*}

\begin{table*}[ht!]
\centering
\resizebox{\linewidth}{!}{%
\begin{tabular}{cccccccccc} 
\hline
\multirow{2}{*}{\textbf{Model}} & \multicolumn{3}{c}{\textcolor[rgb]{0.114,0.11,0.114}{\textbf{Train}}} & \multicolumn{3}{c}{\textcolor[rgb]{0.114,0.11,0.114}{\textbf{Validation}}} & \multicolumn{3}{c}{\textcolor[rgb]{0.114,0.11,0.114}{\textbf{Test}}} \\
 & \textcolor[rgb]{0.114,0.11,0.114}{F1} & \textcolor[rgb]{0.114,0.11,0.114}{precision} & \textcolor[rgb]{0.114,0.11,0.114}{recall} & \textcolor[rgb]{0.114,0.11,0.114}{F1} & \textcolor[rgb]{0.114,0.11,0.114}{precision} & \textcolor[rgb]{0.114,0.11,0.114}{recall} & \textcolor[rgb]{0.114,0.11,0.114}{F1} & \textcolor[rgb]{0.114,0.11,0.114}{precision} & \textcolor[rgb]{0.114,0.11,0.114}{recall} \\ 
\hline
\textcolor[rgb]{0.114,0.11,0.114}{\textbf{BoW}} & \textcolor[rgb]{0.114,0.11,0.114}{1.000} & \textcolor[rgb]{0.114,0.11,0.114}{1.000} & \textcolor[rgb]{0.114,0.11,0.114}{1.000} & \textcolor[rgb]{0.114,0.11,0.114}{0.909} & \textcolor[rgb]{0.114,0.11,0.114}{0.926} & \textcolor[rgb]{0.114,0.11,0.114}{0.897} & \textcolor[rgb]{0.114,0.11,0.114}{0.839} & \textcolor[rgb]{0.114,0.11,0.114}{0.845} & \textcolor[rgb]{0.114,0.11,0.114}{0.838} \\
\textcolor[rgb]{0.114,0.11,0.114}{\textbf{CNN}} & \textcolor[rgb]{0.114,0.11,0.114}{0.998} & \textcolor[rgb]{0.114,0.11,0.114}{0.998} & \textcolor[rgb]{0.114,0.11,0.114}{1.000} & \textcolor[rgb]{0.114,0.11,0.114}{0.920} & \textcolor[rgb]{0.114,0.11,0.114}{0.932} & \textcolor[rgb]{0.114,0.11,0.114}{0.911} & \textcolor[rgb]{0.114,0.11,0.114}{0.883} & \textcolor[rgb]{0.114,0.11,0.114}{0.886} & \textcolor[rgb]{0.114,0.11,0.114}{0.882} \\
\textcolor[rgb]{0.114,0.11,0.114}{\textbf{BioBERT}} & \textcolor[rgb]{0.114,0.11,0.114}{0.960} & \textcolor[rgb]{0.114,0.11,0.114}{0.955} & \textcolor[rgb]{0.114,0.11,0.114}{0.965} & \textcolor[rgb]{0.114,0.11,0.114}{0.946} & \textcolor[rgb]{0.114,0.11,0.114}{0.946} & \textcolor[rgb]{0.114,0.11,0.114}{0.946} & \textcolor[rgb]{0.114,0.11,0.114}{0.914} & \textcolor[rgb]{0.114,0.11,0.114}{0.904} & \textcolor[rgb]{0.114,0.11,0.114}{0.925} \\
\textcolor[rgb]{0.114,0.11,0.114}{\textbf{BioBERT+Lesion Size}} & \textcolor[rgb]{0.114,0.11,0.114}{1.000} & \textcolor[rgb]{0.114,0.11,0.114}{1.000} & \textcolor[rgb]{0.114,0.11,0.114}{1.000} & \textcolor[rgb]{0.114,0.11,0.114}{0.921} & \textcolor[rgb]{0.114,0.11,0.114}{0.921} & \textcolor[rgb]{0.114,0.11,0.114}{0.927} & \textcolor[rgb]{0.114,0.11,0.114}{0.923} & \textcolor[rgb]{0.114,0.11,0.114}{0.920} & \textcolor[rgb]{0.114,0.11,0.114}{0.930} \\
\hline
\end{tabular}
}
\caption{\label{train_val_classification_performance}Model performance for cancer status
classification in training, validation and
test sets.}

\end{table*}

\begin{table*}[!ht]
\centering

\arrayrulecolor{black}
\begin{tabular}{|c|l|} 
\cline{1-1}\arrayrulecolor{black}\cline{2-2}
\multicolumn{1}{|l|}{} & \multicolumn{1}{c|}{\textbf{Hyperparameters}} \\ 
\arrayrulecolor{black}\cline{1-1}\arrayrulecolor{black}\cline{2-2}
\textbf{BoW} & \begin{tabular}[c]{@{}l@{}}TfidfVectorizer:\\\begin{tabular}{@{\labelitemi\hspace{\dimexpr\labelsep+0.5\tabcolsep}}l@{}}Minimum frequency for building vocabulary: 0\textasciitilde{}100\%\\Maximum number of words: 800\textasciitilde{}2500\end{tabular}\\\\SVM:\\\begin{tabular}{@{\labelitemi\hspace{\dimexpr\labelsep+0.5\tabcolsep}}l@{}}Regularization strength C: 0.001\textasciitilde{}1000\\Kernel: \{linear, rbf, poly\}~\end{tabular}\end{tabular} \\ 
\arrayrulecolor{black}\hline
\textbf{CNN} & \begin{tabular}[c]{@{}l@{}}CNN:\\\begin{tabular}{@{\labelitemi\hspace{\dimexpr\labelsep+0.5\tabcolsep}}l@{}}Embedding size: \{64, 96, 128, 200, 256, 512\}\\Size of kernels: \{{[}3], [3, 6], [3, 4, 5], [3, 6, 9], [3, 4, 5, 6]\}\\Number of output channels: \{32, 64, 100, 128, 256\}\end{tabular}\\\\Training:\\\begin{tabular}{@{\labelitemi\hspace{\dimexpr\labelsep+0.5\tabcolsep}}l@{}}Batch size:~ \{16, 32, 50, 64, 128\}\\Learning rate: \{0.001, 0.005, 0.01, 0.05, 0.1\}\\Dropout: 0.01\textasciitilde{}0.5\\Number of epochs: 256\textasciitilde{}2048\\Early stopping patience: 20\textasciitilde{}50\\Weight decay: \{0.0, 0.01, 0.05, 0.1, 0.15\}~\end{tabular}\end{tabular} \\ 
\hline
\textbf{BioBERT} & \begin{tabular}[c]{@{}l@{}}Fine-tuning:\\\begin{tabular}{@{\labelitemi\hspace{\dimexpr\labelsep+0.5\tabcolsep}}l@{}}Batch size: \{4, 8, 10, 12\}\\Learning\_rate: 1.0e-7\textasciitilde{}5e-5\\Dropout: 0.001\textasciitilde{}0.5\\Number of fine-tuning epochs: \{5, 6, 7, 8, 9, 10\}\end{tabular}\end{tabular} \\
\hline
\end{tabular}
\caption{\label{hpo}Searched Hyperparameters for the cancer status classification models. We conducted 100 iterations of hyperparameter search for each of the cancer status classification models via Hyperband-enabled early-stopping \cite{li2017hyperband} Bayesian optimization \cite{snoek2012practical, shahriari2015taking} using  \href{https://wandb.ai/site/sweeps}{Weights \& Biases’ sweeps}. The best hyperparameters were determined using macro-F1 in the validation set. For NER model training, we used batch size=4, learning rate=5e-5, warmup\_steps=2, lower\_case=True, and n\_epoch=10. 
}
\end{table*}

\end{document}